\RequirePackage{fix-cm}
\documentclass[twocolumn]{svjour3}          % twocolumn
\smartqed  % flush right qed marks, e.g. at end of proof
\usepackage{graphicx}
\usepackage{amsmath, amssymb}
\usepackage{float}
\usepackage{epstopdf}
\usepackage{amsmath} 
\usepackage{booktabs}
\begin{document}

\title{The importance of silhouette optimization in 3D shape reconstruction system from multiple object scenes}

%\title{The importance of silhouette optimization for 3D Shape reconstruction in Multi-object scenes}
%\title{An improve approach for 3D Shape reconstruction from multiple-object scenes based on silhouette optimization}
\author{Waqqas-ur-Rehman Butt \and
       Martin Servin %etc.
}
\institute{Waqqas-ur-Rehman Butt \and Martin Servin \at
              Umea University \at              
              90187, SE \at
              \email{ waqqas.butt@umu.se}}          %  \\
\date{Received: date / Accepted: date}
\maketitle

\begin{abstract}

This paper presents a multi stage 3D shape reconstruction system of multiple object scenes by considering the silhouette inconsistencies in shape-from-silhouette \emph{SFS} method. These inconsistencies are common in multiple view images due to object occlusions in different views, segmentation and shadows or reflection due to objects or light directions. These factors raise huge challenges when attempting to construct the 3D shape by using existing approaches which reconstruct only that part of the volume which projects consistently in all the silhouettes, leaving the rest unreconstructed. As a result, final shape are not robust due to multi view objects occlusion and shadows. In this regard, we consider the primary factors affecting reconstruction by analyzing the multiple images and perform pre-processing steps to optimize the silhouettes. Finally, the 3D shape is reconstructed by using the volumetric approach \emph{SFS}. Theory and experimental results show that, the performance of the modified algorithm was efficiently improved, which can improve the accuracy of the reconstructed shape and being robust to errors in the silhouettes, volume and computational inexpensive.	

\end{abstract}

\section{Introduction}
\label{intro}
Three dimensional \emph{3D} shape reconstruction system combines the theories and approaches in the computer vision, which describe the process of ascertaining spatial relationships from data, collected by 3D laser sensors or cameras and modelled the 3D information. The ultimate objective is to understand the real world objects visually by computer. Shape-from-silhouette \emph{SFS} is a 3D reconstruction modality where the available data are the 2D binary projections of the shape in multiple views i.e. the silhouettes \cite{Haro2012}. Usually, the shape is figured as the intersection of the visual cones formed by the silhouettes and their corresponding optical centers, called visual hull \emph{VH}. \emph{SFS} is a fast and simple technique when there are few cameras available and/or multiple objects are present in the scenes. Even though, it is not accurate due to inconsistent silhouettes that are common problems, i.e. occlusion, incomplete shapes, poor lightening , segmentation errors. In these situations, the performance of various multi-view stereo methods \cite{Seitz2006,Furukawa2015} are also reduced.

In this paper, we consider the challenges that are involved to optimize the silhouettes and improve the 3D shape reconstruction process such as shadow filtering, object occlusion from different views and segmentation of largest object in the multiple object scene. 
Firstly, camera calibration is applied to extract the camera parameters. In the next, improved silhouettes are extracted from all views along by applying pre-processing steps. 3D reconstruction is initialized by using space carving \cite{Broadhurst2001}. Finally, the carved voxels are then converted into triangular mesh model to estimate the shape volume. Furthermore, this paper describes a complete, end-to-end system explained in detail in which the essential algorithms are given so that interested reader can easily implement the approach. 

Rest of the paper is organized as follows. Firstly, we describe a survey of literature on the related strands. Section 2 presents the state of art containing, image acquisition system, pre-processing, \emph{SFS}, space carving, 3D voxel model generation and conversion into triangular mesh. The experimental results and evaluation are described in section 3. Finally, the paper is concluded in section 4.
\section{Related work}
\label{Related Work}
%Text with citations \cite{RefB} and \cite{RefJ}.
The object appearance and geometric information can be easily obtain from 2D images. Nevertheless, a single image gives limited geometric information and additional input is required to obtain the veridical 3D model such as multiple images, pre-existing 3D model or a combination thereof  \cite{Su2015}. Perceiving the 3D world is at the heart of successful computer vision applications in robotics, rendering and modelling \cite{Szeliski2010}. For instance, 3D shape synthesis or completion could serve a strong evidence for occlusion reasoning in 2D images \cite{Hsiao2014} and autonomous robots operating in real-world environment \cite{Aleotti2012}. 
In the past, different human assistance approaches for image based 3D shape reconstruction were developed in \cite{Xu2011,Zheng2012,Chen2013,Kholgade2014}. Human assistance was used to segment the image and establish the correspondence between image and models \cite{Xu2011}, fit cuboids and generalized cylinders to objects in images by user \cite{Zheng2012,Chen2013}, user align a stock 3D shape model to a photograph, thus enabling advanced image editing\cite{Kholgade2014}. In these approaches, the partial shape deformation for constraint satisfaction and initial 3D models are essential prerequisite which brings additional user interaction. 

3D shape estimation by using multi-view stereo \cite{Seitz2006,Furukawa2015} and data-driven approaches\cite{Fouhey2013,Su2014} are based on image matching formulation, where the geometric information of an image collection produced 3D models with correspondences through images as described in . The effectiveness of these prior approaches are based on the effective correspondences, views must be close together and matching features must be found in the images, which is a geometry correspondence problem addressed in \cite{Hartley2003}. Moreover, data-driven approaches work only on particular categories and depend on available 3D models, for learning to reason over occlusions.

In order to avoid the limitations of data-driven and multi-view stereo approaches, volumetric scene modelling methods: shape from photo-consistency, shape from shading or motion and shape from silhouettes are proposed in \cite{Seitz1999,Kutulakos2000,Prakash2010,Worthington2001,Yu2007}. The photo-consistency methods requires camera parameters for each view, colour consistency constraints and criteria for consistency checker with specific threshold or from user input. In addition, a model for the object surface reflectance is also required \cite{Mulayim2003}. 
Space carving technique is use to allocate voxels to a 3D object by using the photo-consistency of a specific point with all its matching pixels from the given set of images and carves the 3D object from its background \cite{Kutulakos2000,Prakash2010}. The voxel model produced by space carving is invariably noisy and limited to the visibility of surrounding objects that can be merge into the final 3D shape even they are separated in other views\cite{Kim2006}. 

The shapes from shading and motion approaches \cite{Worthington2001,Yu2007} are effective for object recognition but are less reliant on accurate features correspondences and dependent on prior information about reflectance, shape, viewpoints, and illumination. 

Shape from silhouettes \emph{SFS} techniques \cite{Haro2010,Mulayim2003,Hirano2009,Diaz-Mas2010} deal with the reconstruction of 3D solid models by volume intersection of projected cones, so called visual hull \emph{VH}. All cones have their own peak and axis orientation obtained from camera position and parameters. Some excess volume and enlarge shape is produced in this approximated \emph{VH} due to inconsistent silhouettes and insufficient camera viewing variety. It is very difficult to engrave the additional volume, filling by the concavities, even if infinite number of images is used. The shape from inconsistent silhouette \emph{SFiS} method is presented in \cite{Landabaso2008} to overcome the silhouette inconsistency. It decides the inconsistent silhouette cone intersections that have to be produced so that it can be determined that a voxel is part of the shape and needs to be placed in the \emph{VH}. The unnecessarily enlarge shape is computed due to the re-projection error and occlusion.

The \emph{SFS} methods proposed in \cite{Hirano2009,Mulayim2003,XuandJosephD.EifertandPengchengZhan2006} used a turnable table with fixed cameras to determine the camera parameters corresponding to views and the rotational axis of turntable. However, these methods produced the calibration error due to table motion and required high processing time. The silhouettes methods are proposed in \cite{Franco2009,Haro2012,Hirano2009} are computational extensive which make difficult to parallelise the code and also final shape are not robust due to occlusion, shadows and poor light. Background subtraction $(BS)$ technique to extract the silhouettes in \cite{Hirano2009,Franco2009,Cheung2003}. These methods have some restrictions such as controlled lightening condition and colour information. As a result, the final silhouettes are inconsistent as addressed in \cite{Haro2012}. For an explanation of the unnecessary enlargement of the silhouettes due to shading ia shown in Fig. 3(f). After correction \emph{SFS} reconstruct the shape without an excessive enlargement as shown in Fig. 3(e) .

Silhouette is refined by removal of shadow artefacts in  \cite{6816868,Shoaib,Gong2017,WangShadow}. These methods require prior knowledge about the shape and source of light \cite{6816868}, intensity variation boundary pixels line by user \cite{Gong2017}, $(BS)$ with object contour information by using Gaussian mixture model \emph{GMM} \cite{Shoaib}. Image intensity distributions and moving occlusion boundaries of the object is used to remove the shadow effects as described in \cite{WangShadow}. The application of these algorithms are limited because such information cannot be easily acquired in most scenes, parameterization and also required more processing.

The volume of irregular shape object is measured from 3D reconstructed by mathematical
formulation or counting the number of voxel are proposed in \cite{XuandJosephD.EifertandPengchengZhan2006,Goni2007}. The method \cite{XuandJosephD.EifertandPengchengZhan2006} used \emph{SFS} by single camera along with turntable and then constructed the 3D wire frame model. \emph{RGB} colour ranges from four camera images are used in \cite{Goni2007}, segmentation of object is done by object’s hue surface model and chromaticity similarity, followed by morphological operations to form silhouette. However, these methods are requires the extensive resources for processing \cite{Russ2012} and rotation or mirrors are not suitable methods for the 3D imaging of multiple objects. 

%In this paper, we propose a multistage methodology for 3D shape reconstruction of the largest object in the cluttered scene that is robust to inconsistent silhouette and volume measurement. Firstly, camera calibration is applied to extract the camera parameters. In the next, improved silhouettes are extracted from all views along by applying pre-processing steps. 3D reconstruction is initialized by using space carving \cite{Broadhurst2001}. Finally, the carved voxels are then converted into triangular mesh model to estimate the shape volume. Furthermore, this paper describes a complete, end-to-end system explained in detail in which the essential algorithms are given so that interested reader can easily implement the approach. 
\section{Methodology}
\label{sec:2}
In this section, we describe our approach in order to minimize the errors between silhouettes and projected shapes. The overall diagram of the proposed method is shown in Fig. 1. \begin{figure}[!ht]
\begin{center}
	\includegraphics[scale=.53]{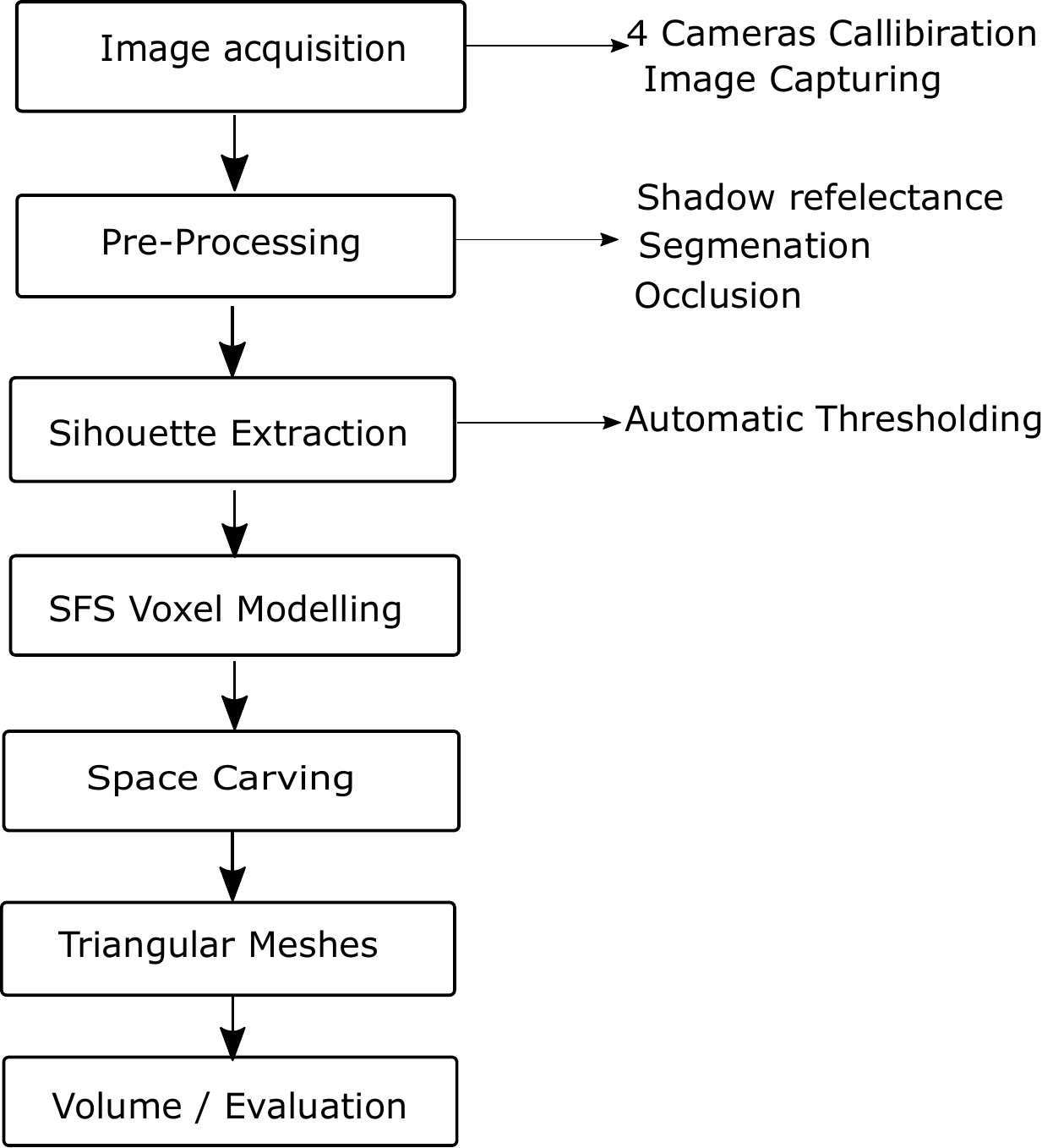}
\caption{System diagram for enhanced 3D shape reconstruction system}
\label{fig:2}        % Give a unique label
\end{center}
\end{figure}
\subsection{Image acquisition system}
We used four fixed cameras in different positions with fixed distance and resolution $(640\times480)$ as shown in Fig.2. A camera captures image from the top of object and three cameras capture image from surrounding object. For instance, if a camera 1 indicates that a voxel is part of the object and camera 2 or 3 close to the former indicates the opposite, then there is an inconsistency that might be solved by a other camera 4 as described in \cite{Diaz-Mas2010}.
The camera parameters are computed by using Matlab calibrator app along with chessboard pattern because it is easy to implement and robust. The strong features of chessboard as dark and bright grids, corners of the grids can be detect and recognize precisely.  These parameters are used to project real world coordinate into image coordinate and eliminate the different effects from an image.
% Block Diagram
\begin{figure}[]
\includegraphics[scale=.05]{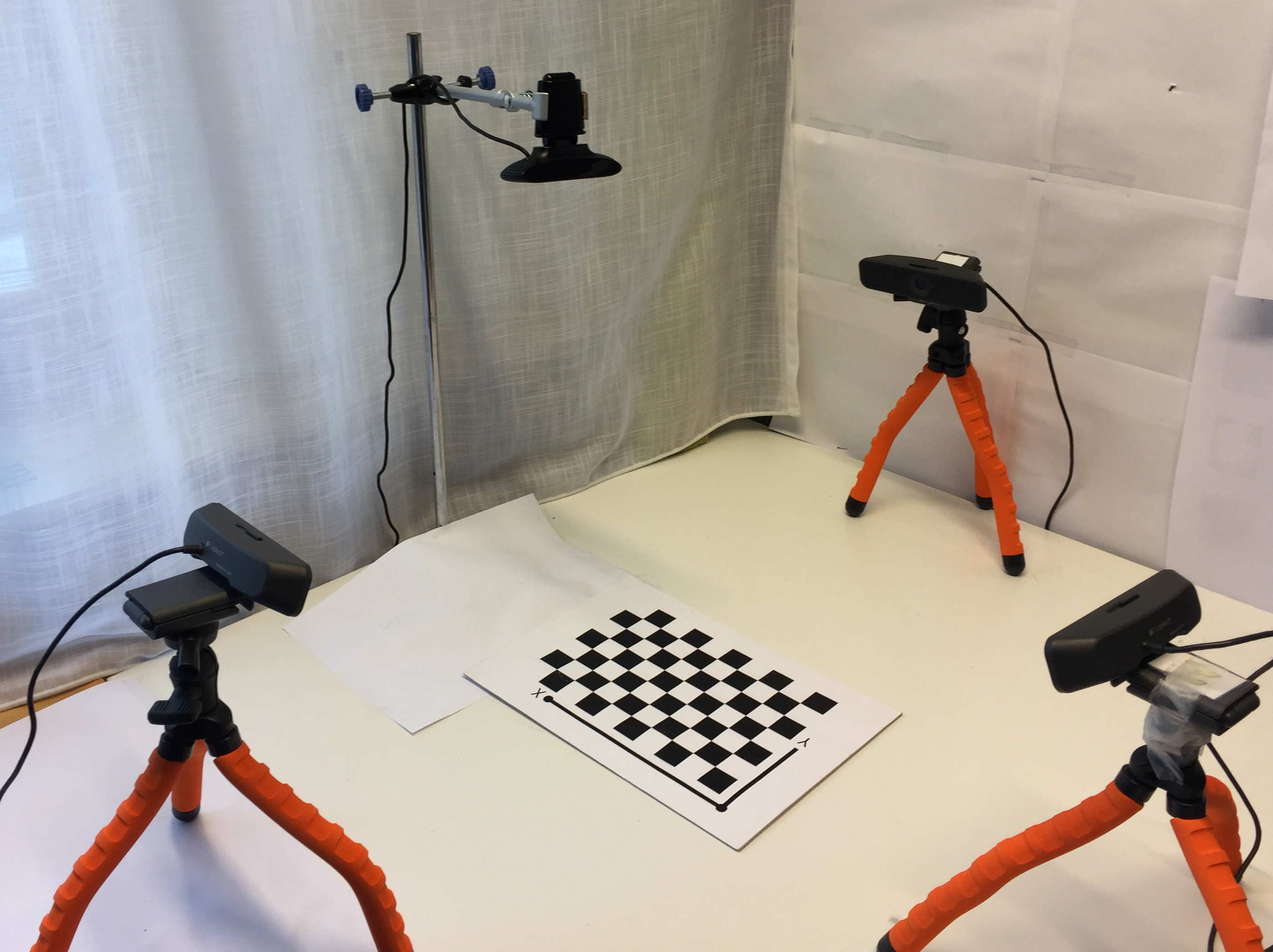}
\caption{Image acquisition system}
\end{figure}
\subsection{Pre-processing}
In this section, we applied pre-processing filters to avoid limitations such as occlusion and shadows from different views as addressed in \cite{Haro2012}. Shaded region are merged with actual object as shown in Fig.3(h), cause the serious problems for accurate silhouette extraction such as enlargement of shape, volume and decrease the reliability of the appearance of the object. At the first stage, all the objects is detected as shown in Fig. 3(e) and then segmented the largest object for simplification of reconstruction process.  
%Previously, some shadow detection and removal methods are proposed in  \cite{6816868,Shoaib,Gong2017,WangShadow}. These methods require prior knowledge about the shape and source of light \cite{6816868}, intensity variation boundary pixels line by user \cite{Gong2017}, $(BS)$ with object contour information by using Gaussian mixture model(GMM) \cite{Shoaib}. Silhouette is refined by removal of shadow artefacts by using image intensity distributions and moving occlusion boundaries of the object as described in \cite{WangShadow}. The application of these algorithms are limited because such information cannot be easily acquired in most scenes, parameterization and also required more processing.
In this contrast, firstly we normalize the pixels intensity values to estimate the brighten edges of the objects and correct the non-uniform illumination. The input image is converted into \texttt{unit8} grayscale and normalize the image by dividing each window size $(3\times3)$ within the image by maximum pixel value that exists within the window. The number of intensity pixels are distributed through the intensity range as shown in Fig. 3(c). In the next, binary mask image is determined by performing the thresholding method. The selection of robust thresholding method is always a challenging task in the computer vision applications. The enhanced Otsu thresholding method \cite{Zhu2009} is used to automatic estimation of the threshold parameters, the resultant image shows the effectiveness of the thresholding of input image as shown in Fig. 3(d). Finally, morphological operation (opening) along with largest connected component analysis (blob analysis) are applied to remove the noise and segmentation of largest object as shown in Fig. 3(e) and 3(f). Fig. 3(h) is displaying the segmented object in red colour in the input image. It can be clearly seen that shaded area has been successfully obsoleted as compare to the method proposed in \cite{WangShadow} as shown in Fig. 3(g).
\begin{figure}
\begin{center}
\includegraphics[scale=.70]{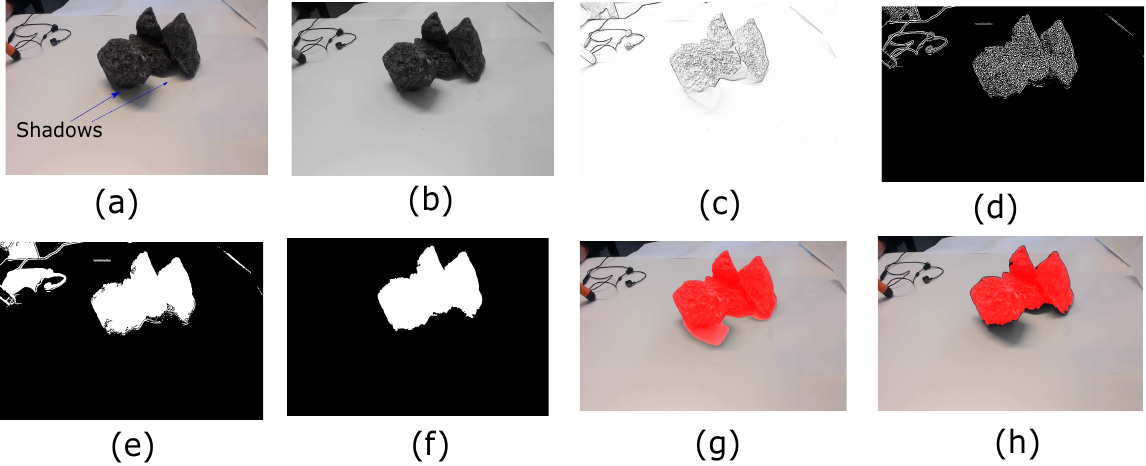}
\caption{Shaded region detection and removal: (a) Input image (b) grayscale image (c) Illuminated image (d) thresholded image  (e) morphological operation (opening)  (f) largest connected component analysis and filling gap (g) result \cite{WangShadow} (h) result (Mask) in red colour}
\label{fig:3}        % Give a unique label
\end{center}
\end{figure}
\subsection{Silhouette extraction}
Traditional \emph{SFS} techniques deal with the reconstruction of 3D solid models by volume intersection from a set of silhouette images as performed in this study. Silhouette is a 2D binary image, extracted from the scene as shown in Fig. 3(f). It is decisive step to obtain accurate the silhouette in complex scene as discussed in previous section (3.2). We extracted the silhouettes from all views by defining the bounding volume \emph{BV} through visual cones that consist of actual object. These cones are created by projection of each silhouette into 3D surface through the correspondent center point of the cameras as shown in Fig. 4.  The intersection of all cones gives the estimated object shape and visual hull\emph{VH} which showed some concave regions that do not appear in silhouette, are not recovered by \emph{SFS}. 
An example scenario is demonstrated in Fig. 4(b), where the silhouette is defined to be the intersection of the visual cones, each cone formed by projecting the silhouette image into the 3D space through the camera center.Fig. 5 is showing the optimized silhouettes from four different views.
\begin{figure}[!ht]
\begin{center}
\includegraphics[scale=.30]{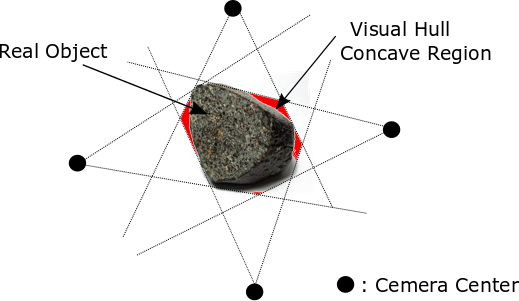}
\caption{Visual hull of the object from center point of cameras by \emph{SFS} method}
\label{fig:4}        % Give a unique label
\end{center}
\end{figure}
\begin{figure}[!ht]
\begin{center}
\includegraphics[scale=.50]{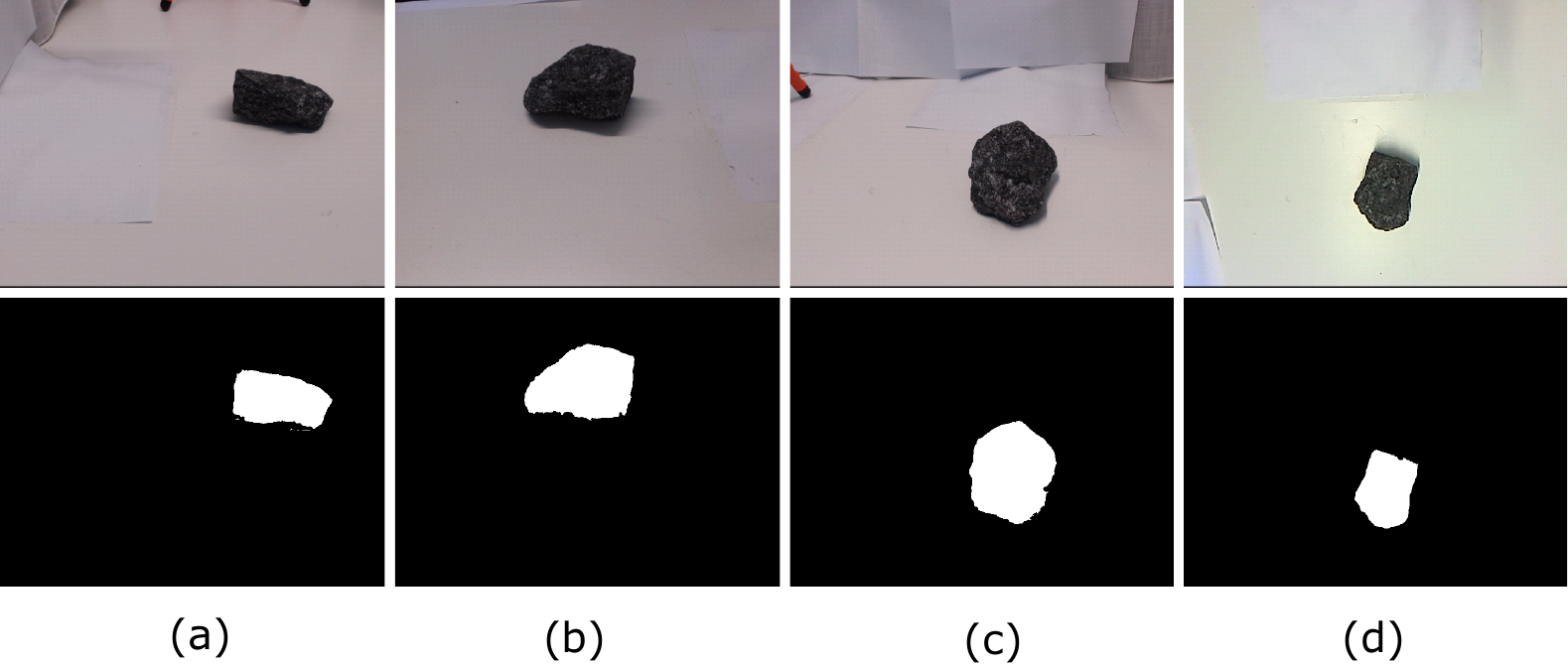}
\caption{Silhouette extraction of by four views a,b,c and d}
\label{fig:5}        % Give a unique label
\end{center}
\end{figure}
\subsection{Voxel modelling} 
A voxel denotes a value on a regular grid in 3D space, also known as volume pixel. After obtaining the \emph{VH}, we then discretize it by dividing into voxels rather than 2D surface. 
The space of interest in $VH$ is divided into voxels $(N1 \times N2 \times N3)$ and projected each voxel onto all the silhouette images by using the correspondence camera parameters (Sec. 3.1). This creates a 3D grid of elements as shown in Fig. 6.
The voxel models is categorised into two groups: if the voxel projection is on all silhouette images classified as inside voxel otherwise outside. The approximation of the \emph{VH}  can be estimated by summation of all the inside voxels and assign scalar values to grid points as follows: 0 to outside, 1 to surface and 2 to inside grid points. The smallest voxel size defines the resolution of the \emph{VH}. 3D grid will be use for carving.
\begin{figure}[!ht]
\begin{center}
\includegraphics[scale=.45]{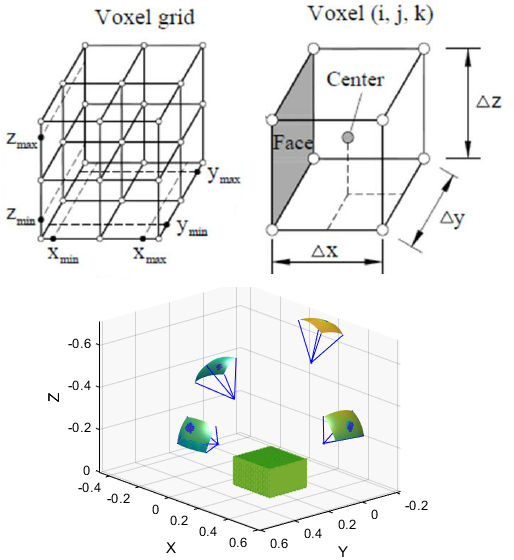}
\caption{Voxel 3D grid with initial bounding volume BV}
\label{fig:6}        % Give a unique label
\end{center}
\end{figure}

\subsection{Space carving}
In the voxel modelling, non-object voxels are obtained due to cavities around the object that do not appear in silhouette. This can lead to incorrect volume and shape. Therefore, space carving method is used to eliminate the non-object voxels. It carves the non-object voxels in hull iteratively by using epipolar geometry and photo consistency through their camera parameters. 
In order to improve the voxel carving process, the photo consistency measure is used at every iteration to decide whether a voxel is object or non-object voxel at given initial bounding volume \emph{BV} as described in \cite{Broadhurst2001}.
%\begin{enumerate}
 %  \item Given the initial bounding volume $BV$
 %  \item Determine the photo consistency of all voxels $V$ in $BV$
 %  \begin{itemize}
 %    \item Project $V$ to all silhouettes in $BV$
%     \item Check the consistency of projected $V$ (pixels wise) within $BV$  to the corresponding optical centers
%   \end{itemize}
%   \item If no non-photo-consistent voxel is found
%   \begin{itemize}
%     \item Terminate
%     \item Otherwise, repeat Step 2.
%   \end{itemize} 
%\end{enumerate}
The final shape is photo consistent with all input images (all views) if all the points are consistent within the $BV$. Furthermore, assign the green colour to the voxels. Fig. 7 is showing the carving results of four camera views. 
\begin{figure}[!ht]
\begin{center}
\includegraphics[scale=.40]{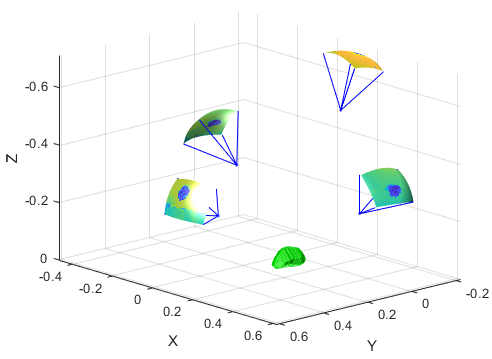}
\caption{Space carving results of four camera views}
\label{fig:7}        % Give a unique label
\end{center}
\end{figure}
\subsection{Voxel to triangular mesh and volume estimation}
A voxel model is confined with 3D grid used for the volume estimation of 3D object as discussed in section (3.4). To improve the accuracy of volume estimation, triangulation method is applied to make sure that all polygons are triangles, called triangular mesh. The marching cube algorithm along with laplacian smoothing method is used for triangulation in \cite{Hirano2009}, but the resultant mesh size is higher as compare to real object. To avoid this inaccuracy, we used the method proposed in \cite{Zhang2001} by defining the basic characteristics of voxel model such as size of block and grid dimension to determine the vertices coordinates $x,y,z$ and edges. Then, calculate the normal $N$ for all edges in normalize vector form which is perpendicular to the corresponding edge and pointing outwards of the mesh. The right-hand rule are applied to obtain the direction of the $N$. A set of triangles is constructed by connecting all the vertices with the origin as shown in Fig. 8.
\begin{figure}[!ht]
\begin{center}
\includegraphics[scale=.08]{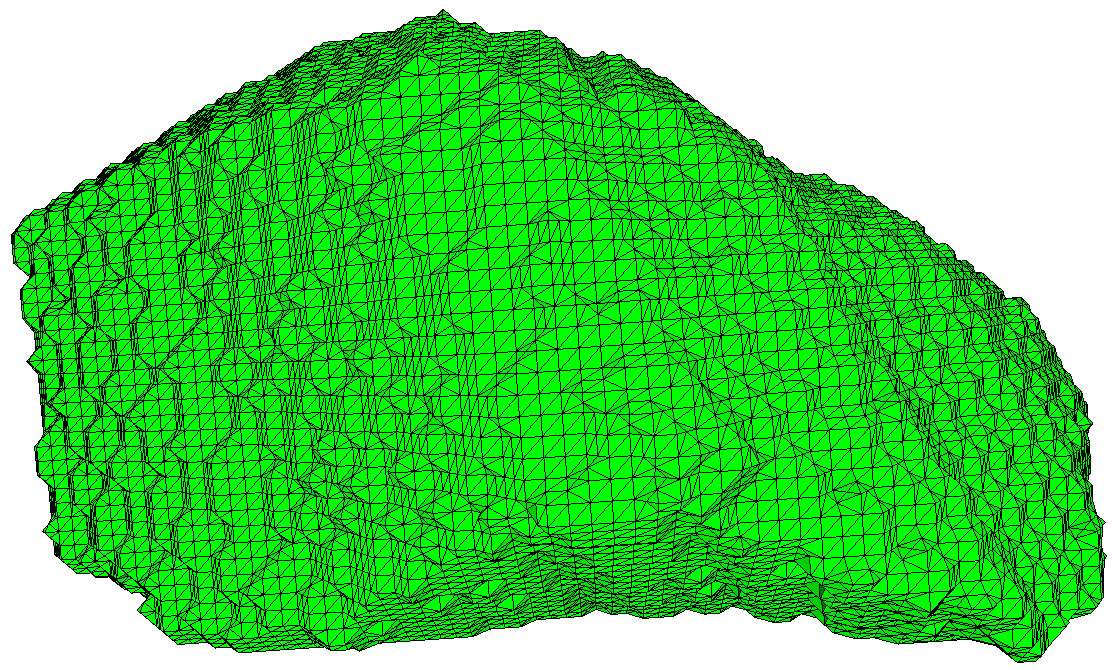}
\caption{Triangular mesh}
\label{fig:8}        % Give a unique label
\end{center}
\end{figure}
The $N$ of the two neighbouring triangles are consistent, if the common edge has different direction. The tetrahedron is obtained by connecting all the vertices of each triangle with the origin and from a tetrahedron. Finally, signed volume is computed by sum of all tetrahedron as described in \cite{Zhang2001}.
\subsection{Real volume measurement}
In this study, our object is rock which has irregular shape, so there is not consistent height or base. To measure the real volume of the object, water displacement method is used and compare the results with trinaglar mesh volume as we calculated in section (3.6). A graduated cylinder is used. The cylinder is filled with distilled water and note the initial water volume. Then, drop the object into cylinder, ensuring that object must be fully submerged in the water and note the new water volume. Finally, real volume of the object is obtained by subtracting the initial volume from the volume of water with object. The position of meniscus was recorded before and after placement of each object in the graduated cylinder with measuring accuracy of $\pm 0.5$ mL. The real volume of the above experiment is mentioned in table 1 as experiment 4.
\section{Experimental results and analysis}
In this section, we present the results of four experiments performed in different conditions such as shadows from light and object reflection, objects occlusion in multi views and segmentation of larger object and compare the results with existing methods. The input images are captured in multi-views as described in section (3.1). The optimized silhouette is obtained by using the automatic threshold value as discussed in Sec. (3.2). The results showed that after silhouette optimization \emph{SFS} method is fast and significant improvement in volume measurement on the traditional \emph{SFS} methods.
Table 1 and 2 show the quantitative comparison, which depend on the results of reconstructed shape volume (compared with real volume) and computational time. The experiment 4 results were obtained from the case used in methodology section(3.3).
\subsection{Experiment 1}
In the first experiment, we use single object with shading effect on different sides of the object as shown in Fig. 9. First column (a) shows the original images captured from four different angles, second column (b) show the results of optimize silhouette by removing the cavities around the object. For a better visualization and understanding, silhouettes are projected in red colour on correspondent original images as shown in Fig. 9c and 9d. We can see that shadow area is successfully discarded from the all the views and which gives the closest 3D shape to the real object as shown in Fig. 9e. Whereas, final shape Fig. 9(f) is bigger in size as shown in Fig. 9c and volume respectively (See Table 1). 
\begin{figure}[!ht]
\begin{center}
\includegraphics[scale=.80]{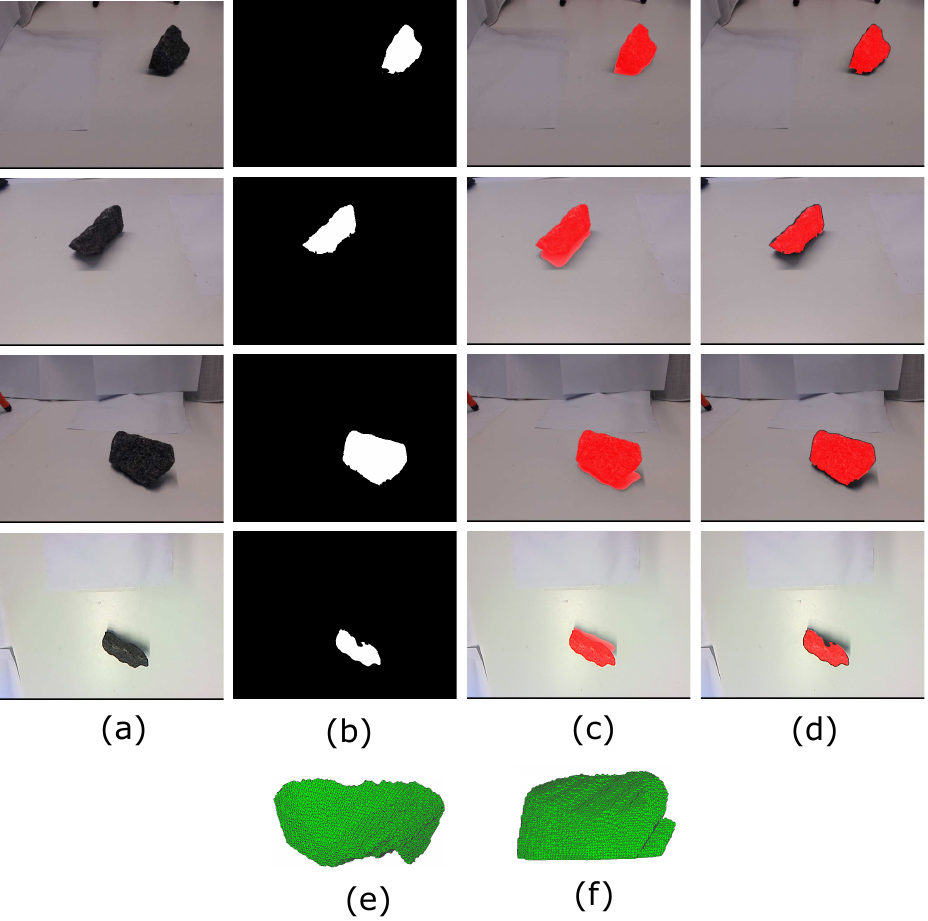}
\caption{Exp 1. 3D shape reconstruction of single object(a) input images (b) optimized silhouette (c) inconsistence silhouette \cite{Hirano2009} (b) optimized silhouette on original images (e) 3D shape after silhouette optimization (f) 3D shape without silhouette optimization}
\label{fig:8}        % Give a unique label
\end{center}
\end{figure}
\subsection{Experiment 2}
In this experiment, reconstruct the shape by using two object that are combined from all views, with complex shading and small gaps. As it can be seen from the Fig. 10(f), the existing method is not robust in the presence of empty area between the objects. This empty part is visible and marked as part of silhouette, which lead to the enlargement of final shape and incorrect volume as shown in Fig. 10(f). In all our experiments, we consider this factor. Comparing the results, we can draw the conclusion that our algorithm can get robust silhouette and shape by successfully removing the shading effects as shown in Fig. 10(e).
\begin{figure}[!ht]
\begin{center}
\includegraphics[scale=.80]{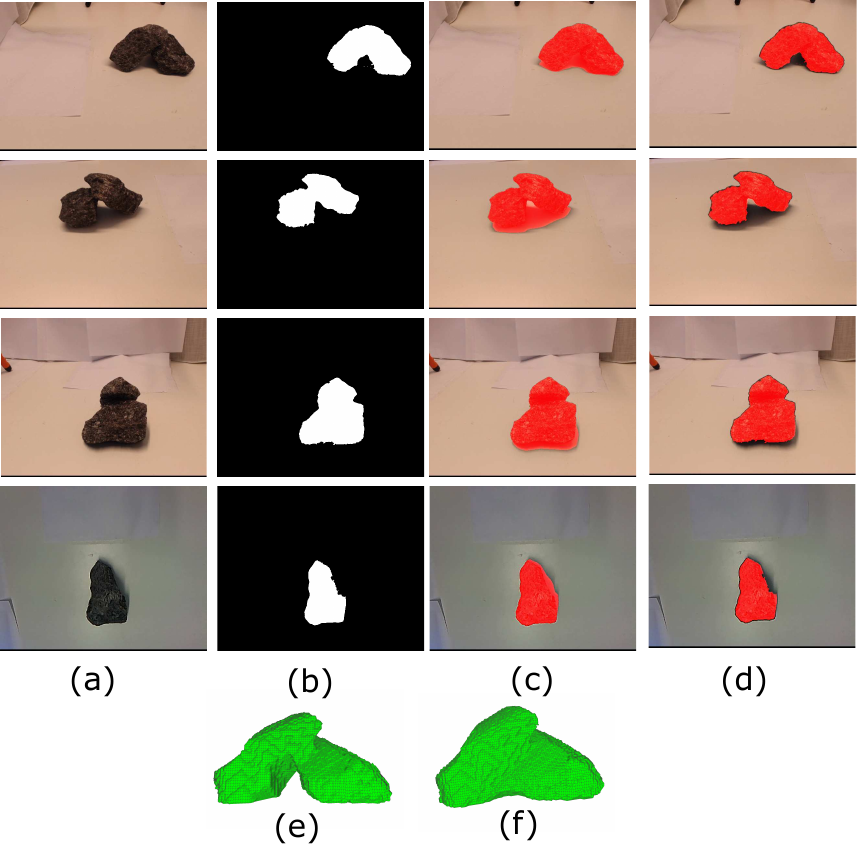}
\caption{Exp 2. 3D reconstruction shape of two combined objects with small gaps and shading t(a) input images (b) optimized silhouette (c) inconsistence silhouette \cite{Hirano2009} (b) optimized silhouette on original images (e) 3D shape after silhouette optimization (f) 3D shape without silhouette optimization}
\label{fig:8}        % Give a unique label
\end{center}
\end{figure}
\subsection{Experiment 3}
The last experiment corresponds to a cluttered scene with multiple objects, complex lighting, objects occlusions in multi-views and colour similarities. We placed a large size object in the center. The optimize silhouette has been extracted without occlusion reasoning in forth camera view as shown in Fig. 11a. In the rest three views, the largest object is partially occluded. As a result, the 3D shape of largest object is reconstructed on the bases of forth camera view by voxel carving as described in Sec. (3.4). On the other hand, inconsistent silhouettes are obtained from all views, which leads to incorrect shape and volume as shown in Fig. 11(f).
\begin{figure}[!ht]
\begin{center}
\includegraphics[scale=.70]{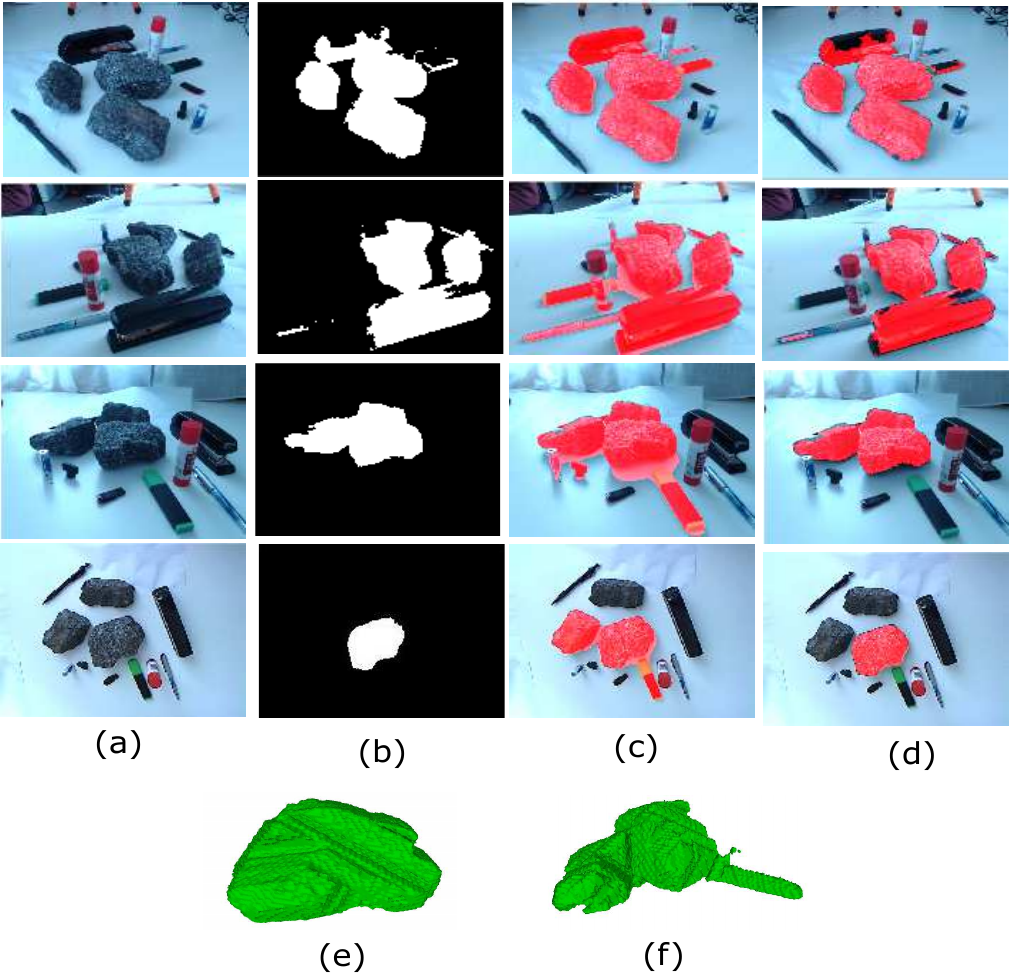}
\caption{Exp 3. 3D reconstruction shape of larger object in complex scene t(a) input images (b) optimized silhouette (c) inconsistence silhouette \cite{Hirano2009} (b) optimized silhouette on original images (e) 3D shape after silhouette optimization (f) 3D shape without silhouette optimization}
\label{fig:8}        % Give a unique label
\end{center}
\end{figure}
\subsection{Performance evaluation}
We evaluate the performance of \emph{SFS} method by considering silhouette inconsistencies in series of experiments.  All the experiments have been successfully carve the convex hull of the visible object and reconstructed the shape, especially in experiment 2 and 3, where the object characteristics are more influent. Despite, it has been possible to segment of largest object and volume measurement. The quality of the reconstructed shape in terms of volume are generally accurate. However, the final volume of the shape is not 100 \% accurate but is close to the real volume as mentioned in Table. 1. As a base line, we compared the results with real volume by experimental triangular mesh volume as described in sec. (3.6) and (3.7). 
The relative percent error $RE$ and precision are determined by equation 1 and 2. The relative uncertainty is used to quantitatively express the precision of a measurement.
\begin{equation}
RE= \frac{X-X_{0}}{X_{0}}\times 100 
\end{equation}
\begin{equation}
Precision= \frac{RE}{X_{0}}\times 100
\end{equation}
Here $X$ and $X_{0}$ are real and experimental values of the volume measurement. 
Table 1 shows a quantitative comparison of the results. As it can be seen from table 1, the enhance \emph{SFS} method is better at precision, the average precision in four experiments is 1.39 \%. In comparison, the average precision existing model is 4.98\%, considerably higher than proposed approach.
\begin{table}
	\centering
	\caption{Comparison of Volume in cubic centimeter $cm^{3}$ with measuring accuracy $\pm 0.5$ mL}
	\label{my-label}
	\begin{tabular}{lllll}
		%\hline
		\textbf{Exp.} & \textbf{Algorithm} & \textbf{Exp.} & \textbf{Real} & \textbf{Uncertainty}  \\  
		$ $ & $  $ & \textbf{Volume}  $  $ & \textbf{volume} & \textbf{ (\%)} \\ \hline% <--
		1            & \begin{tabular}[c]{@{}l@{}}\cite{Hirano2009}\\ Proposed\end{tabular} & \begin{tabular}[c]{@{}l@{}}290.7\\ 258.9\end{tabular} & 240  & \begin{tabular}[c]{@{}l@{}}7.27\\ 3.04\end{tabular}            \\ \hline
		2            & \begin{tabular}[c]{@{}l@{}}\cite{Hirano2009}\\ Proposed\end{tabular} & \begin{tabular}[c]{@{}l@{}}712.8\\ 559.2\end{tabular} & 565   & \begin{tabular}[c]{@{}l@{}}3.67\\ 0.18\end{tabular}          \\ \hline
		3            & \begin{tabular}[c]{@{}l@{}}\cite{Hirano2009}\\ Proposed\end{tabular} & \begin{tabular}[c]{@{}l@{}}1214.8\\ 694.1\end{tabular} & 720 & \begin{tabular}[c]{@{}l@{}}5.66\\ 0.52\end{tabular}             \\ \hline
		4 & \begin{tabular}[c]{@{}l@{}}\cite{Hirano2009}\\ Proposed\end{tabular} & \begin{tabular}[c]{@{}l@{}}237.4\\ 211.5\end{tabular}  & 220   & \begin{tabular}[c]{@{}l@{}}3.33\\ 1.83\end{tabular} 
		\\ \hline
	\end{tabular}
\end{table}

The processing time required to construct the shape depends upon the complexity of scene and computational resources. The automatic silhouette creation step has a positive impact on the 3D reconstruction. 
The shape constructed by using the Laplacian smoothing \cite{Hirano2009} is generally required more processing.
%The shape constructed by \cite{Hirano2009} is generally  required more processing time due to the conversion process from the voxel model to the polygonal model using the Laplacian smoothing
In order to reduce the computational time, triangular mesh modeling is used rather Laplacian smoothing. The processing time is calculated by using a PC with a Core i7 2.66 GHz with 8GB RAM. Table 2 show the comparison of the reconstruction time for \emph{SFS} after silhouette optimization is projected for fast reconstruction. We achieve this goal: the triangular meshes are adequate and avoids the non-concave parts in the final shape.
\begin{table}[]
\centering
\caption{Comparison of processing time in seconds}
\label{my-label}
\begin{tabular}{llll} \hline
\textbf{Exp.} & \textbf{\cite{Franco2009}} & \textbf{\cite{Hirano2009}} & \textbf{proposed}  \\ \hline
1             & 65.33            & 60.45             & 47.2              \\\hline
2             & 42.56            & 39.58             & 31.16             \\\hline
3             & 42.8             & 40.64             & 35.66             \\\hline
4             & 85.44            & 99.65             & 72.44             \\ \hline
\end{tabular}
\end{table}
\subsection{Advantages}
In order to take the advantages of new digital image processing techniques, it is possible to enhance the 3D reconstruction shapes in different manners. (1) To extract better shape by minimizing the effects that silhouette have over the reconstructed Shape.(2) To recover errors in the silhouettes from informed decisions made at the volume level. Contrarily, previous \emph{SFS} approaches \cite{Xu2011,Zheng2012,Chen2013,Kholgade2014,Mulayim2003} does not only fail to recover errors in the silhouettes but it worsens the silhouettes by propagating 2D misses from one view to the others. (3) \emph{VH} based techniques in that they can operate with no initialisation, that is, no a-priori knowledge about the scene. (4) Triangular mesh model is generated to measure the volume of shape, which can be used in commercial CAD/CAE/CAM systems.
%After correction, the result indicates that the \emph{SFS} method is quicker in processing and can successfully carved the concave areas, handle highly reflective objects and segmentation of largest object in the corresponding input images. This may be attributed to the fact that over segmentation that it may produce is not usually a problem, and the flexibility this adds improves reliability.

\section{Conclusion}
\label{Sec:conclusion}
%\textbf{
%This paper proposes new methodology for reconstruction of 3D shape from cluttered scenes in multiple views. These cluttered scenes pose several challenges such as object occlusions in views, shadows due to light or object or surface reflection and segmentation. Many traditional $SFS$ approaches produced a smooth 3D shape but the final shape is bigger in size and also computational extensive. In order to overwhelm these challenges, we present a complete image based 3D shape reconstruction system from multiple 2D images based on silhouette enhancement. There are three key factors are added in this study, silhouette optimization by considering the occlusions, shading/reflection due to light or object and unsupervised segmentation of largest object. Triangular mesh model is  to measure the volume of object. The experimental results compared with existing methods, demonstrate the robustness of proposed method in shape, volume and lower computational complexity. As regards with precision, the proposed method gives the closest result to the real volume. The average precision in volume measurement is 1.39 \%, shows the effectiveness of the proposed system. Moreover, the final shape can be used in different field of applications such as surveillance and autonomous systems. As a future research, we want to compare the results by using high-resolution cameras and silhouette classification by convolution neural network (CNN).  

In this paper, we have considered the silhouette inconsistencies as usually found in practical scenarios by using \emph{SFS} method. There are three key factors are added to optimize silhouette (1) object occlusions in views, (2)shading/reflection due to light or object and (3) unsupervised segmentation of largest object. 
%The experimental results compared with existing methods, demonstrate the robustness of modified \emph{SFS} method in shape, volume and lower computational complexity. 
After correction, the result indicates that the \emph{SFS} method is quicker in processing and can successfully carved the concave areas, handle highly reflective objects and segmentation of largest object in the corresponding input images. 
%This may be attributed to the fact that over segmentation that it may produce is not usually a problem, and the flexibility this adds improves reliability. 
As regards with precision, the average precision in volume measurement is 1.39 \% by proposed method which gives the closest result to the real volume and highly practical for constructing 3D shape of real objects. Moreover, the final shape can be used in different field of applications such as Digital 3-D objects are used in many areas films, computer games, educational/learning tools and interactive mapping, surveillance and autonomous systems.
%As a future research, we want to compare the results by using high-resolution cameras and silhouette classification by convolution neural network (CNN).
\section{Compliance with ethical standards}
This research was in partly supported by the Kempe Foundations (SMK-1644.1) to Ume\aa\ University.

\textbf{Conflict of interest : }The authors declare that they have no conflict of interest.

\textbf{Ethical approval : }This article does not contain any studies with human participants or animals performed by any of the authors.

%}
%\begin{acknowledgements}
%\textbf{
%This research was in partly supported by the Kempe Foundations (SMK-1644.1) to Ume\aa\ %University. 
%}
%\end{acknowledgements}

\bibliographystyle{unsrt}

\end{document}